\documentclass[runningheads]{llncs}

\usepackage{eccv}

\usepackage{eccvabbrv}

\usepackage{graphicx}
\usepackage{booktabs}

\usepackage[accsupp]{axessibility}  

\usepackage{xspace} 
\usepackage{wrapfig, lipsum, booktabs}

\newcommand{\diffhdr}{DiffHDR\xspace}
\usepackage{hyperref}

\usepackage{orcidlink}

\makeatletter
\renewcommand\paragraph{\@startsection{paragraph}{4}{\z@}%
{-8\p@ \@plus -3\p@ \@minus -3\p@}%
{-0.5em \@plus -0.22em \@minus -0.1em}%
{\normalfont\normalsize\itshape}
}
\makeatother

\makeatletter
\renewcommand\subsubsection{\@startsection{subsubsection}{3}{\z@}%
{-4\p@ \@plus -1\p@ \@minus -1\p@}%
{-0.5em \@plus -0.22em \@minus -0.1em}%
{\normalfont\normalsize\bfseries\boldmath}}
\makeatother

\begin{document}

\title{
DiffHDR: Re-Exposing LDR Videos with Video Diffusion Models \\ 
}

\titlerunning{DiffHDR}

\author{Zhengming Yu\inst{1,2} \and Li Ma\inst{2} \and Mingming He\inst{2} \and Leo Isikdogan\inst{3} \and Yuancheng Xu\inst{3} \and Dmitriy Smirnov\inst{3} \and Pablo Salamanca\inst{3} \and Dao Mi\inst{3} \and Pablo Delgado\inst{3} \and \\Ning Yu\inst{3} \and Julien Philip\inst{2} \and Xin Li\inst{1} \and Wenping Wang\inst{1} \and Paul Debevec\inst{3}}

\authorrunning{Z. Yu, L. Ma, M. He et al.}

\institute{Texas A\&M University \and
Eyeline Labs \and
Netflix}

\maketitle

\begin{abstract}
Most digital videos are stored in 8-bit low dynamic range (LDR) formats, where much of the original high dynamic range (HDR) scene radiance is lost due to saturation and quantization. 
This loss of highlight and shadow detail precludes mapping accurate luminance to HDR displays and limits meaningful re-exposure in post-production workflows. Although techniques have been proposed to convert LDR images to HDR through dynamic range expansion, they struggle to restore realistic detail in the over- and underexposed regions.
To address this, we present \diffhdr, a framework that formulates LDR-to-HDR conversion as a generative radiance inpainting task within the latent space of a video diffusion model. 
By operating in Log-Gamma color space, \diffhdr leverages spatio-temporal generative priors from a pretrained video diffusion model to synthesize plausible HDR radiance in over- and underexposed regions while recovering the continuous scene radiance of the quantized pixels. 
Our framework further enables controllable LDR-to-HDR video conversion guided by text prompts or reference images. 
To address the scarcity of paired HDR video data, we develop a pipeline that synthesizes high-quality HDR video training data from static HDRI maps. 
Extensive experiments demonstrate that \diffhdr significantly outperforms state-of-the-art approaches in radiance fidelity and temporal stability, producing realistic HDR videos with considerable latitude for re-exposure.
\end{abstract}

\begin{figure*}[!t]
    \centering
    \includegraphics[width=0.95\linewidth]{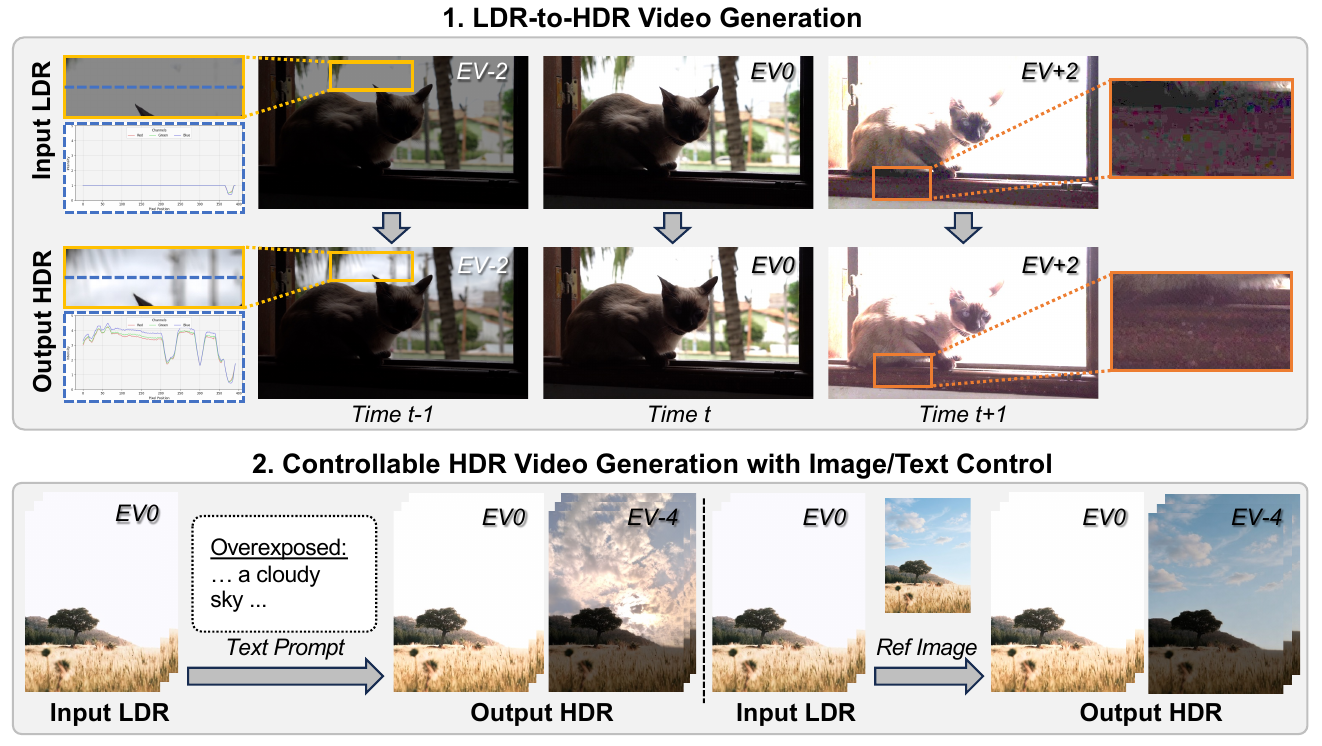}
    \vspace{-2ex}
    \caption{\textbf{DiffHDR} reconstructs lost radiance to convert LDR videos into faithful HDR while maintaining temporal coherence (Top). \diffhdr further enables controllable HDR synthesis guided by text prompts or reference images, facilitating realistic hallucination of saturated regions (Bottom).}
    \vspace{-3ex}
    \label{fig:Teaser}
\end{figure*}

\section{Introduction}
\label{sec:intro}

High dynamic range (HDR) video captures a wide range of scene luminance, preserving intricate details across both deep shadows and extreme highlights. This capability not only enables more faithful visual reproduction on HDR displays, but also provides crucial flexibility in post-production workflows such as color grading, tone mapping, and re-exposure.
Despite these benefits, the vast majority of digital video is confined to low dynamic range (LDR) formats, including almost the entirety of video produced using generative models. This LDR-centric ecosystem persists because LDR remains the most portable format for consumer hardware, while true HDR acquisition typically requires high-end cameras or complex multi-exposure techniques that are often impractical for everyday use. Furthermore, recent advanced video generative models~\cite{wan2025wan,blattmann2023stable} are mostly trained on large-scale 8-bit LDR datasets, further entrenching these dynamic range limitations.
Therefore, there is a critical need for effective LDR-to-HDR conversion methods which can hallucinate missing scene radiance and unlock the inherent potential of HDR within existing LDR content.

Existing LDR-to-HDR approaches can be broadly categorized into two groups. The first class reconstructs HDR content in a multi-exposure fusion setting \cite{patchbased,hdrvideo,deephdrvideo,kalantari2017deep}, which requires a sophisticated capture set-up and is not practical for the single LDR video setting.
The other generates HDR images from a single LDR input, typically using a feed-forward deep neural network \cite{eilertsen2017hdr,liu2020single,yu2021luminance,santos2020single,Guo_2022_ACCV,marnerides2018expandnet,sdrtohdrtv,generativehdrrecon}. 
Due to the limited model capacity and their deterministic pixel-to-pixel translation formulation, these methods often struggle to synthesize photorealistic content in clipped regions. 
Fundamentally, LDR-to-HDR conversion is a one-to-many problem because of the appearance ambiguity in over- and underexposed regions. 
This naturally motivates the use of generative models. Training such a generative model remains challenging due to the lack of large-scale, high-quality HDR video datasets.
Therefore, a more practical solution is to leverage the strong priors of video models pretrained on large-scale LDR video.
However, this is nontrivial, as models trained on LDR videos do not natively support HDR content due to the fundamental distribution mismatch between LDR and HDR videos.

To address these challenges, we propose \diffhdr, the first video diffusion-based framework for generative reconstruction of HDR videos from a single LDR video.
The key enabler of our approach is a deceptively simple observation that HDR videos, when processed with carefully designed tone-mapping curves, can be aligned with the manifold of a video VAE trained on LDR videos. Specifically, we introduce a Log-Gamma color mapping which compresses high dynamic range content into the operational range of the pretrained video VAE, enabling HDR videos to be encoded and decoded without any finetuning. To overcome data scarcity, we develop a curated generation pipeline which leverages high-quality panoramic HDRI maps from Polyhaven~\cite{polyhaven} to synthesize a diverse HDR video dataset. 
Despite finetuning solely on synthetic videos derived from static HDRIs, our framework generalizes robustly to real-world videos by leveraging the strong priors of the pretrained video model. To address information loss in clipped regions, we employ luminance-based masks to guide both the generative process and a context-focused cross-attention module. By incorporating context-focused prompting or reference images, this module facilitates controllable reconstruction in over- and underexposed areas, utilizing spatio-temporal cues to hallucinate physically plausible details (Fig.~\ref{fig:Teaser}). Our main contributions are as follows:
\begin{enumerate}
\item We introduce \diffhdr, the first video diffusion framework for LDR-to-HDR reconstruction, along with a curation pipeline which synthesizes high-quality HDR video training data from static HDRIs.
\item We introduce a Log-Gamma color mapping, enabling HDR generation within pretrained latent spaces while preserving the backbone's generative priors and temporal consistency without any VAE finetuning.
\item We design exposure-aware control mechanisms with luminance-based mask detection, context-focused prompting, and context-focused cross-attention to enhance controllable generation in over- and underexposed regions.
\item \diffhdr achieves state-of-the-art performance on synthetic and in-the-wild benchmarks, significantly outperforming prior methods in radiance fidelity and temporal stability while enabling downstream applications such as text- and image-guided HDR video editing.
\end{enumerate}

\section{Related Work}
\label{sec:related_work}

\subsection{Multi-Exposure Fusion}

HDR videos can be captured directly using specialized hardware such as beam splitters or multi-sensor camera systems~\cite{hdrproductionsystem,multisensorHDR}. While effective, these solutions are typically expensive and impractical for widespread deployment.
As a result, recovering HDR content from standard LDR images has become an attractive alternative.

The classical paradigm reconstructs HDR images from multiple photographs captured with different exposure settings~\cite{debevec1997hdr}, commonly known as multi-exposure fusion.
However, capturing multi-exposure image sequences often leads to spatial misalignment due to camera motion or dynamic scene content, making naive fusion prone to artifacts.
Early methods addressed this issue by explicitly aligning multi-exposure images using global or local registration techniques~\cite{kang2003hdrvideo,sen2012robust,localnonrigid,Hu_2013_CVPR}.
Learning-based approaches have since shown advantages over explicit alignment pipelines.
Convolutional neural network (CNN)–based methods generate HDR images directly from misaligned multi-exposure inputs by implicitly handling alignment or fixing misalignment artifacts during reconstruction~\cite{kalantari2017deep,wu2018deep,xiong2021hierarchical,niu2021hdr,multiscale,kong2024safnet}.
Transformer-based architectures further improve performance by modeling long-range dependencies~\cite{yan2019attention,yan2020deep,ye2021progressive,chen2023improving,liu2022ghost,yan2023smae,song2022selective,tel2023alignment}.

The multi-exposure paradigm has also been extended to HDR video reconstruction, where different exposure settings are temporally interleaved across frames to provide complementary information~\cite{patchbased,hdrvideo,deephdrvideo,kalantari2017deep}.
In addition to compensating for inter-frame motion, HDR video methods must also enforce temporal consistency to avoid flickering and other temporal artifacts~\cite{xu2024hdrflow,Chen_2021_ICCV,Chung_2023_ICCV}. Despite their success, multi-exposure fusion methods typically require specialized acquisition setups and are inapplicable to single-exposure LDR inputs.

\subsection{HDR from a Single Image}

While multi-exposure fusion focuses on combining information from multiple LDR images, a complementary line of work aims to generate HDR content from a single LDR input~\cite{eilertsen2017hdr}. 
This can be achieved by explicitly estimating an inverse tone-mapping function~\cite{liu2020single}. 
Another class of methods directly regresses HDR outputs from LDR images using neural networks~\cite{eilertsen2017hdr,yu2021luminance,santos2020single,Guo_2022_ACCV,marnerides2018expandnet,sdrtohdrtv,generativehdrrecon}. 
Some increase dynamic range in intermediate representations, such as gain maps~\cite{liao2025learning,meng2025ultraled} or intrinsic components like shading maps~\cite{intrinsichdr}. 
An alternative strategy predicts multiple virtual LDR images at different exposure levels from a single input, which can then be fused to produce HDR content~\cite{endo2017deep,zhang2023revisiting,Le_2023_WACV,lee2018deep,meng2025ultraled}.

Single-image HDR generation relaxes the capture requirements but introduces a fundamental challenge, where the lost information in overexposed or underexposed regions needs to be re-synthesized.
Several methods explicitly incorporate inpainting modules to hallucinate missing details in saturated regions~\cite{liu2020single,generativehdrrecon,goswami2024semantic}. 
However, when using limited-capacity generative models, the synthesized content often lacks realism or fine details.

\subsection{Generative HDR}

Advances in generative modeling, including GANs~\cite{goodfellow2020generative,karras2019style,karras2020analyzing,karras2021alias,chan2022efficient,trevithick2023real,sun2023next3d,jiang2023nerffacelighting,yu2025gaia,arjovsky2017wasserstein,chan2021pi} and diffusion models~\cite{dhariwal2021diffusion,rombach2022high,mei2025lux,he2024diffrelight,yu2024surf,wang2024disentangled,zhang2025spgen,huang2025vchain,xu2025virtually,yang2024cogvideox,HaCohen2024LTXVideo,opensora,wang2025pdt,wang2023360,zhang2025uniser,agarwal2025cosmos,zhu2023taming}, have shown strong priors for image and video generation.
Some approaches learn the mapping from LDR images to HDR using only LDR videos, without requiring HDR supervision~\cite{whatcanbelearned}. 
Similarly, GlowGAN~\cite{wang2023glowgan} enables GAN-based HDR image generation by learning from the distribution of LDR content.

Diffusion models, in particular, have demonstrated strong capability in generating photorealistic image and video, and have been widely applied to tasks such as controllable generation~\cite{zhang2023controlnet,mou2024t2i,wan2025wan,jiang2025vace,burgert2025go,ju2024brushnet,gu2025diffusion}, editing~\cite{meng2022sdedit,jiang2025vace}, inpainting \cite{lugmayr2022repaint,adiya2024omnipainter}, and restoration \cite{saharia2023image,li2022srdiff}.
These strengths have motivated their adoption in generative HDR creation. Hu \etal~\cite{Hu_2024_CVPR} employ diffusion models to reduce ghosting artifacts in multi-exposure fusion.
UltraFusion~\cite{chen2025ultrafusion} formulates exposure fusion as a guided inpainting task, using a latent diffusion model to hallucinate missing information in overexposed regions with guidance from underexposed inputs.
Bracket Diffusion~\cite{bemana2024exposure} enables pretrained diffusion models in LDR to generate HDR outputs through multiple diffusion passes under different exposure conditions.
HDR-V-Diff~\cite{diffusionpromoted} introduces a latent diffusion model specifically designed for HDR video generation.
Guan \etal~\cite{guan2025hdr} fine-tune diffusion models to jointly generate gain maps and LDR images for HDR generation, while LEDiff~\cite{wang2025lediff} performs HDR generation via latent-space fusion.
Concurrently, X2HDR~\cite{wu2026x2hdr} reuses a pretrained variational autoencoder (VAE) for LDR images by compressing HDR content into the PU21 color space, enabling HDR reconstruction within an LDR-oriented latent representation. However, leveraging pretrained video diffusion priors for controllable HDR generation remains largely unexplored.

\begin{figure*}[t]
    \centering
    \includegraphics[width=1.0\linewidth]{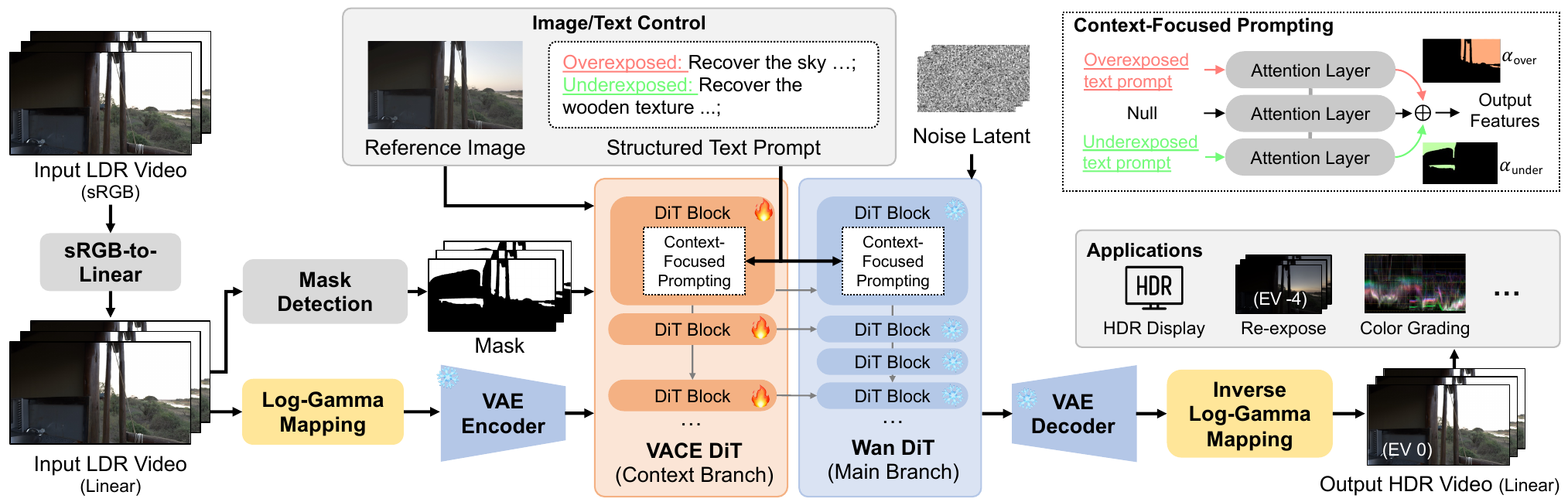}
    \vspace{-2ex}
    \caption{\textbf{Framework of DiffHDR.}
Given an input LDR video, we first detect its clipped regions and map it into the proposed Log-Gamma color space. A finetuned video diffusion model reconstructs missing radiance in over- and underexposed regions. A mask detector and context-focused prompting module support controllable detail synthesis. The final output HDR video supports faithful re-exposure, accurate reproduction on HDR displays, and flexible post-production workflows. }
    \vspace{-3ex}
    \label{fig:pipeline}
\end{figure*}

\section{Method}
\label{sec:method}

We adopt a latent video diffusion framework to achieve controllable LDR-to-HDR conversion. The overall pipeline is illustrated in Fig.~\ref{fig:pipeline}. Due to the lack of existing HDR video data, we first construct a curated HDR video dataset using a data generation pipeline based on static HDRIs (Sec.~\ref{sec:data}). To fully leverage the pretrained video VAE, we introduce a Log-Gamma mapping that compresses HDR value into a bounded range (Sec.~\ref{sec:lg_encoding}). 
The input LDR video is first mapped to the Log-Gamma color space, encoded into latent space using the video VAE.
A finetuned latent video diffusion model then reconstructs plausible radiance in saturated and noisy regions (Sec.~\ref{sec:diffusion}). We adopt VACE~\cite{jiang2025vace}, a video-to-video latent diffusion framework, as our backbone. To enhance controllability, we incorporate structured text prompts and reference-image-based control signals that explicitly guide detail synthesis in over- and underexposed regions (Sec.~\ref{sec:control}).

\subsection{HDR Video Dataset Curation}
\label{sec:data}
\subsubsection{Dataset Curation Pipeline.} 
Training video diffusion models for HDR video reconstruction requires paired LDR–HDR video data.
While LDR video can be synthesized from HDR video, publicly available HDR video data with high fidelity and adequate dynamic range remains limited.
Therefore, we construct a curated HDR video dataset using a rendering-based data generation pipeline built upon 16K resolution HDRIs from Polyhaven~\cite{polyhaven}.

For each HDRI, we place the camera at the origin and set the HDRI as the skybox.
We render short video sequences in Blender using multiple predefined camera configurations. 
Specifically, we design three motion patterns for diverse dynamic range distributions: (1) highlight-focused zoom-in/out sequences emphasizing saturated regions, (2) shadow-focused zoom-in/out sequences emphasizing underexposed areas, and 
(3) camera rotation sequences introducing pseudo dynamics.
For highlight-focused and shadow-focused sequences, we first identify the brightest and darkest pixels in the HDRI, respectively, and orient the camera toward the corresponding areas. 
For zoom-in/out, the start and end focal lengths are randomly sampled between $(18, 30)$ mm and $(50, 70)$mm. For rotation sequences, we rotate the camera around the vertical axis by 120° per segment, obtaining 3 segments to cover the full 360° from each HDRI.

We render videos in linear color space (i.e. Rec.709) from about 800 HDRIs. The resulting dataset includes approximately 5400 HDR video sequences, each containing 81 frames, across diverse illumination environments. 
These sequences provide the temporally consistent HDR supervision that is essential for learning radiance reconstruction and re-exposure within the video diffusion framework.

\subsubsection{Data Augmentation Strategies.}
Given a rendered HDR video, we synthesize its LDR video by simulating the LDR video formation process, including exposure shift, heteroscedastic camera noise, quantization, and clipping.
\vspace{-1ex}

\paragraph{Exposure shift.}
We randomly sample an exposure offset $\Delta \in [-2, 2]$ stops and scale the video in linear space by a factor of $2^{\Delta}$.
\vspace{-1ex}

\paragraph{Camera noise.}
To simulate realistic sensor noise, we follow CBDNet~\cite{guo2019toward} and model camera noise as a heteroscedastic Gaussian process whose variance depends on the signal intensity. Specifically, the noise is formulated as:
\begin{equation}
\mathbf{n}_t(L_t)=\sqrt{L_t\sigma_s^2+\sigma_c^2}\,\boldsymbol{\epsilon}_t,
\end{equation}
where $L_t$ denotes the input pixel intensity in linear space. $\sigma_s$ and $\sigma_c$ are the signal-dependent component and stationary noise, sampled from $(0, 8.5\times10^{-4})$ and $(0, 1.5\times10^{-5})$, respectively. $\boldsymbol{\epsilon}_t \sim \mathcal{N}(0,\mathbf{I})$ is a standard Gaussian noise field.

Unlike CBDNet, which samples noise independently for each image, we extend the model to videos by introducing temporal correlation in the underlying Gaussian noise. Specifically, we share $(\sigma_s,\sigma_c)$ across all frames and model $\boldsymbol{\epsilon}_t$ using an AR(1) (first-order autoregressive) process~\cite{hamilton2020time}:
\begin{equation}
\boldsymbol{\epsilon}_t = \rho \boldsymbol{\epsilon}_{t-1} + \sqrt{1-\rho^2}\mathbf{u}_t,
\end{equation}
where $\mathbf{u}_t \sim \mathcal{N}(0,\mathbf{I})$ and $\rho$ controls the temporal correlation strength. When $\rho=0$, the noise reduces to independent sampling per frame. In our experiments, we set $\rho=0.5$.
\vspace{-1ex}
\paragraph{Quantization and clipping.}
To produce the final LDR inputs, we convert the HDR video to sRGB, clip values to $[0,1]$, and quantize to 8-bit precision.

\subsection{Log-Gamma Color Mapping}
\label{sec:lg_encoding}

The video VAE is a core component in latent video diffusion models~\cite{wan2025wan}. 
However, as shown in Fig.~\ref{fig:ablation_encode}, a VAE pretrained on LDR data fails to accurately encode and decode HDR content, as the pixel values can far exceed the standard $[0, 1]$ range of LDR signals.
While it is possible to finetune the VAE to support HDR content \cite{wang2025lediff,diffusionpromoted}, this approach is hindered by the lack of large-scale, high-quality HDR video datasets. Furthermore, modifying the VAE architecture or weights shifts the learned latent space, potentially disrupting the generative priors of the pretrained model.
Instead of adapting the VAE itself, we introduce a transformation that maps HDR radiance into a representation compatible with the VAE's pretrained domain.
Specifically, we formulate this as a color-mapping function.
Inspired by $\mu$-law tone mapping~\cite{yan2019attention, guan2024diffusion} and perceptual gamma compression in imaging pipelines, we propose a Log-Gamma color mapping defined as:
\vspace{-2pt}
\begin{equation}
\mathcal{T}(x)
=
\left(
\frac{\log\!\left(1 + \gamma x\right)}
{\log\!\left(1 + \gamma M\right)}
\right)^{\frac{1}{\gamma}},
\label{eq:lg_encoding}
\end{equation}
where $x$ denotes the linear HDR radiance, $M$ is the maximum representable radiance, and $\gamma$ regulates compression.
The logarithmic component compresses high dynamic range radiance, while aligning the radiance distribution with natural LDR statistics, ensuring direct compatibility with the pretrained VAE.

\subsection{Diffusion-Based LDR-to-HDR Conversion}
\label{sec:diffusion}

\subsubsection{Preliminary.}
Our model builds upon VACE~\cite{jiang2025vace}, a diffusion-based video-to-video framework for video editing. VACE introduces a Video Condition Unit (VCU) that integrates text prompts, context frames, and masks into a unified conditioning interface. These inputs are encoded into latent tokens via a video VAE and processed by a DiT-based backbone to model spatiotemporal dependencies under the flow matching framework~\cite{lipman2022flow, liu2022flow}.

\subsubsection{Model Architecture of \diffhdr.}
As shown in Fig.~\ref{fig:pipeline}, the input LDR video is first linearized and mapped to the Log-Gamma color space, and then encoded into a latent representation using the VAE encoder. This LDR latent is fed into the context branch to condition the denoising process of the main branch. 
We additionally compute an exposure mask indicating the over- and underexposed regions, guiding the model toward areas that require detail hallucination. 
Starting from a random noise latent, the main branch iteratively denoises to produce the final HDR latent, which is subsequently decoded and inverse Log-Gamma mapped to linear space, yielding the final HDR video suitable for downstream applications.
To enable controllable hallucination in clipped regions, we condition the model on both text prompts and reference images, as detailed in Sec.~\ref{sec:control}.

\subsubsection{Fine-tuning Strategy.}
To preserve the pretrained generative prior of VACE, we freeze the backbone parameters and fine-tune only the DiT blocks via LoRA adapters. 
Specifically, rank-32 LoRA layers are inserted into the attention and feed-forward layers of the DiT blocks. 
This parameter-efficient adaptation ensures stable training while reducing overfitting to the HDR dataset.

\subsubsection{Training Objective.}
The training objective follows a standard rectified flow-matching formulation~\cite{liu2022flow}.
Specifically, given a HDR video sample in latent representation $\mathbf{x}_1$ and a Gaussian noise $\mathbf{x}_0 \sim \mathcal{N}(0, \mathbf{I})$, we sample a timestep $t \in [0,1]$ and construct an intermediate latent via linear interpolation:
\begin{equation}
\mathbf{x}_t = t \mathbf{x}_1 + (1 - t) \mathbf{x}_0.
\label{eq:flow_xt}
\end{equation}
The final objective is defined as:
\begin{equation}
\mathcal{L} = \mathbb{E}_{\mathbf{x}_0, \mathbf{x}_1, t}
\left\|
u_\Theta(\mathbf{x}_t, t, \mathbf{c}) - (\mathbf{x}_1 - \mathbf{x}_0)
\right\|_2^2,
\label{eq:flow_loss}
\end{equation}
where $u_\Theta$ indicates the video DiT and $\mathbf{c}$ denotes the conditioning signals including the LDR input, exposure masks, and optional text or image prompts.

\subsection{Controllable HDR Video Reconstruction}
\label{sec:control}
\subsubsection{Luminance-Based Mask Detection.}

We construct a luminance-based mask detection to identify over- and underexposed regions in LDR videos. The input sRGB frames are first linearized using the inverse sRGB transfer function, and luminance is computed following the Rec.709 standard. 
Over- and underexposed regions are detected by thresholding luminance values: pixels with luminance greater than $\tau_{high}$ are considered overexposed, while those below $\tau_{low}$ are treated as underexposed. 
We set $\tau_{high}=0.95$ and $\tau_{low}=0.05$.

To further improve temporal stability, we perform per-pixel exponential moving average (EMA) smoothing:
\begin{equation}
\tilde{M}_t = \alpha M_t + (1-\alpha)\tilde{M}_{t-1},
\label{eq:temporal_mask}
\end{equation}
where $\alpha$ controls the smoothing strength, $M_t$ denotes the mask detected at time $t$, and $\tilde{M}_t$ is the temporally smoothed mask. 
We set $\alpha = 0.7$. 
This temporal aggregation suppresses frame-wise fluctuations and improves mask consistency for video diffusion conditioning.

\begin{figure*}[!t]
    \centering
    \includegraphics[width=\linewidth]{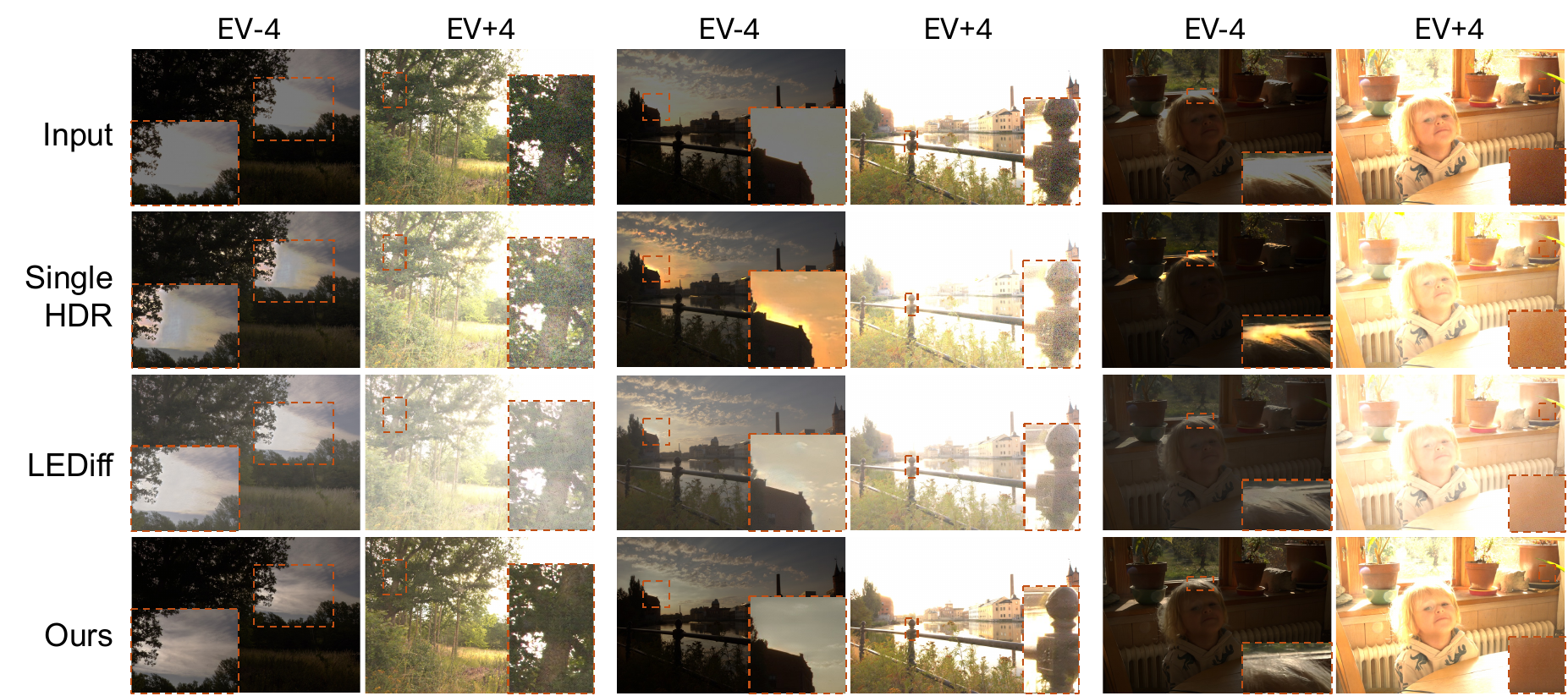}
    \vspace{-3ex}
    \caption{\textbf{Qualitative comparison on the SI-HDR dataset.} 
Results are shown under multiple re-exposure levels to assess highlight restoration, shadow recovery, and radiance consistency across methods. Zoom in for detailed comparison.}
    \label{fig:qual_sihdr}
    \vspace{-3ex}
\end{figure*}
\label{sec:experiments}

\subsubsection{Context-Focused Prompting.}
Our key idea is to design a context-focused captioning format that explicitly grounds the visual semantics of regions with distinct exposure characteristics. Unlike standard prompts that provide a single holistic description of the scene, our prompts follow a structured format:
\texttt{[overexposed: <description>]; [underexposed: <description>]}.
This formulation disentangles the semantic guidance for saturated highlights and shadowed regions, enabling region-aware conditioning.

Inspired by classifier-free guidance~\cite{ho2022classifier}, which manipulates the global denoising trajectory using conditional and unconditional prompts, we introduce a context-focused cross-attention mechanism that operates locally inside the DiT cross-attention blocks. Importantly, this modification is applied exclusively at inference, preserving pretrained weights and training objectives.
Specifically, our CFA module is applied at every cross-attention layer for both the VACE context branch and main branch. Let $\mathbf{x}$ denote the current token features and $\mathbf{c}$, $\mathbf{c}_{\text{over}}$, and $\mathbf{c}_{\text{under}}$ indicate the unconditional embedding, the overexposed text prompt, and the underexposed text prompt, respectively.
The output of the cross-attention layer is
\begin{equation}
\mathbf{r}_{\text{base}} = \mathrm{CA}(\mathbf{x}, \mathbf{c})\text{,} \quad
\mathbf{r}_{\text{over}} = \mathrm{CA}(\mathbf{x}, \mathbf{c}_{\text{over}})\text{, and }
\mathbf{r}_{\text{under}} = \mathrm{CA}(\mathbf{x}, \mathbf{c}_{\text{under}}),
\end{equation}

where $\mathrm{CA}(\cdot)$ denotes the cross-attention operator. 
Given the corresponding spatial masks $\mathbf{M}_{\text{over}}$ and $\mathbf{M}_{\text{under}}$, we then refine the model output using a mask-guided routing mechanism:
\begin{equation}
\mathbf{r} = 
\mathbf{r}_{\text{base}} 
+ \alpha_{\text{over}} \, \mathbf{M}_{\text{over}} \odot (\mathbf{r}_{\text{over}} - \mathbf{r}_{\text{base}})
+ \alpha_{\text{under}} \, \mathbf{M}_{\text{under}} \odot (\mathbf{r}_{\text{under}} - \mathbf{r}_{\text{base}}),
\label{eq:cfa}
\end{equation}

where $\alpha_{\text{over}}$ and $\alpha_{\text{under}}$ control the strength of region-specific modulation, and $\odot$ denotes element-wise multiplication. 

This design preserves the global semantic structure from the base prompt while selectively steering the generation in over- and underexposed regions. Because the modification only alters cross-attention residuals at inference, it is fully compatible with trained DiT models and does not require retraining.

\begin{figure*}[t]
    \centering
    \includegraphics[width=\linewidth]{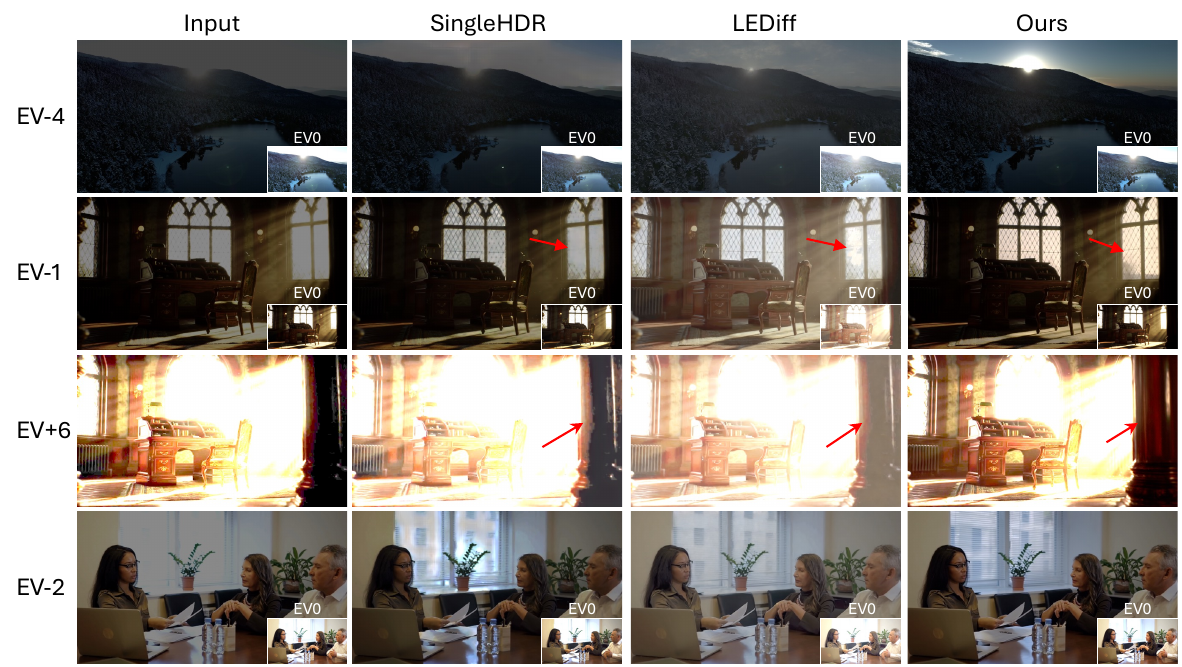}
    \vspace{-2ex}
    \caption{\textbf{Qualitative comparison on in-the-wild video dataset.} Results are shown under multiple re-exposure levels across frames to assess highlight restoration, shadow recovery, and radiance consistency across methods. 
Zoom in for detailed comparison.}
    \vspace{-3ex}
    \label{fig:qual_video}
\end{figure*}

\subsubsection{Reference Image-Based Conditioning.}
Text prompts provide simple and abstract control for guiding the synthesis of new details. However, in some cases more fine-grained control is required. To this end, we allow the user to optionally provide a reference image that specifies detailed appearance cues in LDR. 
Such reference images can be generated using existing image editing models.
To condition the generation process, we follow the VACE architecture and encode the reference image using the VAE. The encoded reference is then concatenated along the temporal dimension to inject the reference signal into the model.

\section{Experiments}

\begin{wraptable}{r}{5cm}
\vspace{-6ex}
\caption{Quantitative comparison on the SI-HDR dataset.}
\label{tab:quan_sihdr}
\centering
\resizebox{0.41\textwidth}{!}{
\begin{tabular}{lccc}
\toprule
Method & HDR-VDP3 $\uparrow$ & PU21-PIQE $\downarrow$ & FID $\downarrow$ \\
\midrule
HDRCNN    & 6.82 & 24.30 & 19.26 \\
MaskHDR   & 6.87 & 24.00 & 19.78 \\
SingleHDR & \textbf{7.37} & 26.64 & 27.55 \\
LEDiff    & 6.56 & 22.71 & 25.98 \\
\textbf{Ours} & 6.98 & \textbf{19.37} & \textbf{18.68} \\
\bottomrule
\end{tabular}
}
\vspace{-4ex}
\end{wraptable}

We evaluate our method across diverse datasets, including SI-HDR dataset~\cite{hanji2022comparison}, the Cinematic Video dataset~\cite{froehlich2014creating}, 50 held-out videos from our Polyhaven-based synthetic dataset (excluded from training). In addition, we collect 50 in-the-wild videos from Pexels~\cite{pexels}, and 10 videos generated using Veo2~\cite{team2023gemini}.

For the SI-HDR dataset, following LEDiff~\cite{wang2025lediff}, we adopt HDR-VDP3~\cite{mantiuk2023hdr}, PU21-PIQE~\cite{hanji2022comparison}, and FID~\cite{heusel2017gans} as evaluation metrics. 
For video benchmarks, we use FovVideoVDP~\cite{mantiuk2021fovvideovdp} as a reference-based HDR video metric, and adopt DOVER~\cite{wu2023exploring}, CLIPIQA~\cite{wang2023exploring}, and MUSIQ~\cite{ke2021musiq} as non-reference perceptual quality metrics, following FlashVSR~\cite{zhuang2025flashvsr}.
These metrics effectively evaluate the spatial and temporal quality of the reconstructed HDR.

We compare \diffhdr against state-of-the-art LDR-to-HDR methods. In addition, we conduct comprehensive ablation studies to validate the effectiveness of each proposed component and demonstrate further applications of our framework. Additional results are provided in the supplementary material.

\subsection{Implementation Details}
We build our framework upon the pretrained video diffusion model Wan-2.1-VACE-14B~\cite{jiang2025vace} and adopt the corresponding Wan-2.1-VAE~\cite{wan2025wan} with a spatiotemporal compression ratio of \(4 \times 8 \times 8\). Our model is trained at a spatiotemporal resolution of \(33 \times 1280 \times 720\). For LoRA adaptation, we insert rank-32 LoRA modules into the DiT blocks while freezing the backbone parameters. The model is trained using the AdamW~\cite{loshchilov2017decoupled} optimizer with a constant learning rate of \(1 \times 10^{-4}\) for 10,000 steps. 
Training is performed on 8 NVIDIA A100 GPUs with mixed precision setting.
Since BF16 precision can introduce banding artifacts in HDR decoding due to its limited precision, we use BF16 for finetuning the DiT and FP32 for the VAE to preserve tonal continuity.

\begin{table}[t]
\centering
\vspace{-2ex}
\caption{Quantitative comparison on Cinematic Video and synthetic datasets.}
\vspace{-2ex}
\label{tab:quan_cinematic}
\resizebox{\linewidth}{!}{
\begin{tabular}{lcccccccc}
\toprule
& \multicolumn{4}{c}{\textbf{Cinematic Video Dataset}} 
& \multicolumn{4}{c}{\textbf{Polyhaven Synthetic Video Dataset}} \\
\cmidrule(lr){2-5} \cmidrule(lr){6-9}
Method 
& FOVVDP $\uparrow$ & DOVER $\uparrow$ & MUSIQ $\uparrow$ & CLIPIQA $\uparrow$
& FOVVDP $\uparrow$ & DOVER $\uparrow$ & MUSIQ $\uparrow$ & CLIPIQA $\uparrow$ \\
\midrule
SingleHDR 
& 6.56 & 0.77 & 51.79 & 0.33
& 7.48 & 0.66 & 57.60 & 0.46 \\

LEDiff    
& 3.75 & 0.63 & 47.90 & 0.28
& 4.68 & 0.59 & 58.92 & 0.46 \\

\textbf{Ours} 
& \textbf{6.89} & \textbf{0.81} & \textbf{58.38} & \textbf{0.41}
& \textbf{7.65} & \textbf{0.68} & \textbf{60.02} & \textbf{0.50} \\
\bottomrule
\end{tabular}
}
\end{table}

\begin{table}[t]
\vspace{-2ex}
\centering
\caption{Quantitative comparison on in-the-wild and Veo2 video datasets.}
\vspace{-2ex}
\label{tab:quan_wild}
\resizebox{0.8\textwidth}{!}{
\begin{tabular}{lcccccc}
\toprule
& \multicolumn{3}{c}{\textbf{In-the-wild Video Dataset}} 
& \multicolumn{3}{c}{\textbf{Veo2 Video Dataset}} \\
\cmidrule(lr){2-4} \cmidrule(lr){5-7}
Method 
& DOVER $\uparrow$ & MUSIQ $\uparrow$ & CLIPIQA $\uparrow$
& DOVER $\uparrow$ & MUSIQ $\uparrow$ & CLIPIQA $\uparrow$ \\
\midrule
SingleHDR 
& 0.71 & 53.21 & 0.46 
& 0.59 & 43.78 & 0.30 \\

LEDiff    
& 0.61 & 53.68 & 0.42 
& 0.53 & 41.41 & 0.29 \\

\textbf{Ours} 
& \textbf{0.74} & \textbf{55.79} & \textbf{0.48} 
& \textbf{0.61} & \textbf{46.06} & \textbf{0.34} \\
\bottomrule
\end{tabular}
}
\vspace{-5ex}
\end{table}

\subsection{Comparisons with State-of-The-Art}
\subsubsection{Quantitative Evaluations.}
We quantitatively compare \diffhdr with state-of-the-art LDR-to-HDR methods on both image and video benchmarks. For all LDR-based perceptual metrics (i.e., FID, PU21-PIQE, MUSIQ, CLIPIQA, and DOVER), we uniformly apply Reinhard tone mapping~\cite{reinhard2005dynamic} to convert HDR outputs to LDR space to ensure fair comparison across methods. 
Although not trained on image data, our method achieves the best performance in PU21-PIQE and FID, and ranks second on HDR-VDP3 as shown in Tab.~\ref{tab:quan_sihdr}. 
The slightly lower HDR-VDP3 score is likely because our method generatively inpaints plausible details in clipped regions that may not be pixel-wise identical to the ground truth and are therefore penalized by this metric.
Nevertheless, these results indicate the superior perceptual quality of our method. 
On video datasets, \diffhdr consistently achieves the best performance on both the Cinematic Video dataset and the Polyhaven synthetic video dataset across reference-based and non-reference metrics as shown in Tab.~\ref{tab:quan_cinematic}. Furthermore, on the in-the-wild and Veo2-generated video datasets, our method exhibits the strongest generalization capability, outperforming prior approaches across all reported metrics as shown in Tab~\ref{tab:quan_wild}. 
These results indicate that \diffhdr produces temporally coherent and visually stable HDR videos, enabling robust HDR reconstruction and re-exposure in dynamic real-world scenes.

\begin{figure*}[!t]
    \centering
    \includegraphics[width=\linewidth]{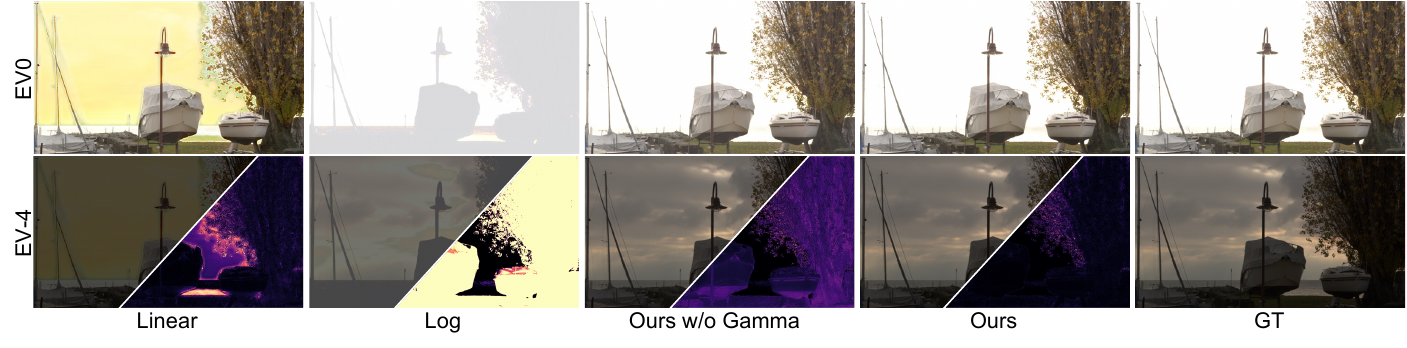}
    \vspace{-5ex}
\caption{\textbf{Comparison of different color mappings.} 
We compare different color mapping methods by feeding the mapped images into the VAE and evaluating the reconstruction quality. We also visualize per-pixel error maps, where brighter regions indicate larger reconstruction errors. Our method achieves the best performance.}
    \vspace{-4ex}
    \label{fig:ablation_encode}
\end{figure*}

\subsubsection{Qualitative Evaluations.}
We present qualitative comparisons on the SI-HDR dataset (Fig.~\ref{fig:qual_sihdr}) and in-the-wild videos (Fig.~\ref{fig:qual_video}). To avoid potential bias introduced by different tone-mapping operators, we directly compare HDR outputs under multiple exposure levels, allowing consistent evaluation of recovered radiance and dynamic range. As shown in Fig.~\ref{fig:qual_sihdr}, \diffhdr effectively restores fine details in severely saturated sky regions and generalizes well to challenging high-intensity structures such as overexposed hair strands. In shadow regions, our method suppresses noise while recovering structural details. In contrast, LEDiff~\cite{wang2025lediff} and SingleHDR~\cite{liu2020single} introduce visible artifacts in saturated areas and struggle to remove camera noise in dark regions. For in-the-wild videos (Fig.~\ref{fig:qual_video}), \diffhdr successfully reconstructs the radiance of the sun with wider dynamic range in the first example, while preserving surrounding high-frequency details. It also restores overexposed window regions and underexposed pillars with improved dynamic range and structural fidelity. LEDiff can approximate the sun’s shape but produces limited dynamic range and flattened highlights. SingleHDR fails to recover accurate structures in saturated regions, with noticeable artifacts. Moreover, both LEDiff and SingleHDR suffer from temporal inconsistencies in high-intensity areas, whereas \diffhdr maintains temporally stable reconstruction.

\subsubsection{Ablation Studies.}
We conduct ablation studies on the Polyhaven synthetic dataset to validate the effectiveness of the proposed components.

\begin{figure*}[!t]
    \centering
    \includegraphics[width=\linewidth]{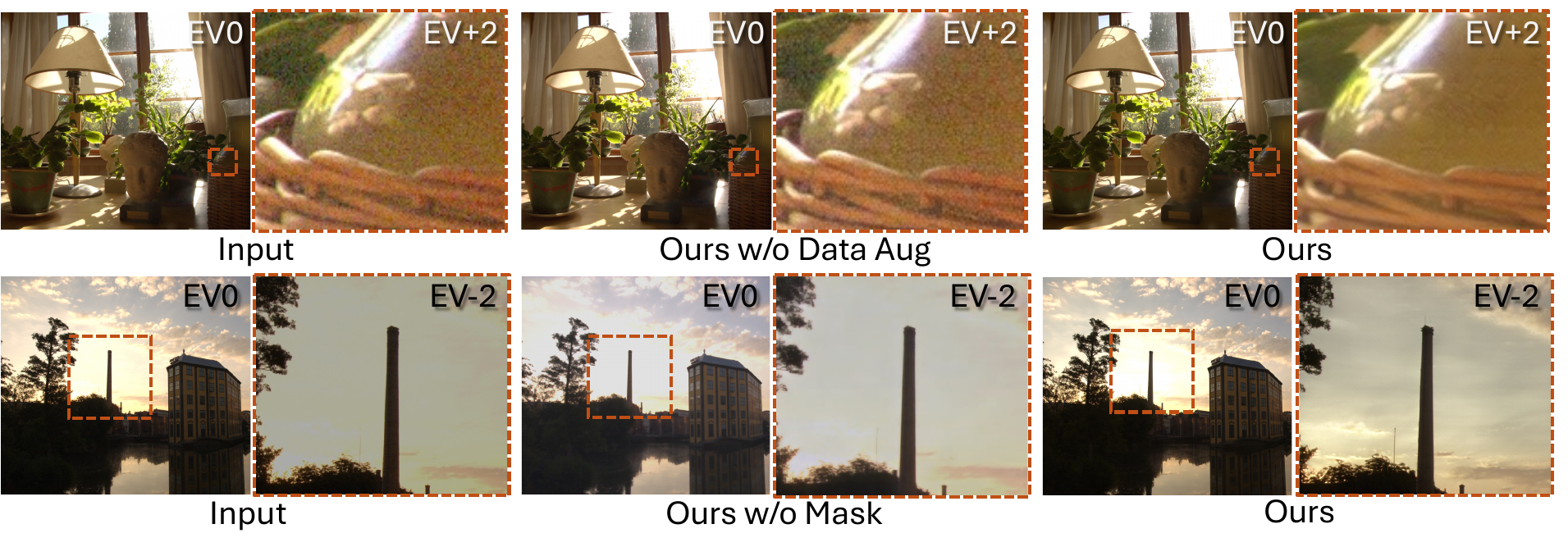}
    \vspace{-5ex}
    \caption{\textbf{Ablation study on data augmentation strategy and mask detection.} We compare our methods without the data augmentation training and the mask detection module.}
    \vspace{-3ex}
    \label{fig:ablation_aug_mask}
\end{figure*}

\paragraph{Effect of Log-Gamma mapping.}
To evaluate the proposed Log-Gamma color mapping, we compare four encoding strategies applied before the VAE encoder and decoder: (1) directly encoding linear HDR values (Linear), (2) a pure logarithmic mapping (Log) used in LEDiff's finetuning, (3) our mapping without gamma compression, $\mathcal{T}'(x)=\frac{\log(1+x)}{\log(1+M)}$, and (4) our full Log-Gamma mapping.

\begin{wraptable}{r}{5cm}
\vspace{-7.5ex}
\centering
\caption{Quantitative comparison of different mapping strategies.}
\label{tab:ablation_encode}
\resizebox{0.41\textwidth}{!}{
\begin{tabular}{lccc}
\toprule
Method & PSNR $\uparrow$ & SSIM $\uparrow$ & LPIPS $\downarrow$ \\
\midrule
Linear          & 22.16  & 0.74 & 0.28 \\
Log             & 14.61 & 0.74 & 0.57 \\
Ours w/o gamma  & 25.38 & 0.75 & 0.34 \\
\textbf{Ours}   & \textbf{32.86} & \textbf{0.86} & \textbf{0.15} \\
\bottomrule
\end{tabular}
}
\vspace{-6ex}
\end{wraptable}

We assess reconstruction quality both quantitatively and qualitatively. As shown in Fig.~\ref{fig:ablation_encode}, both Linear and Log mappings fail to faithfully recover color and structural details of the ground-truth HDR inputs. The mapping without gamma compression introduces noticeable artifacts, particularly around high-contrast edges. In contrast, our full Log-Gamma mapping accurately reconstructs fine details and preserves color consistency.

\begin{figure*}[!t]
    \centering
    \includegraphics[width=\linewidth]{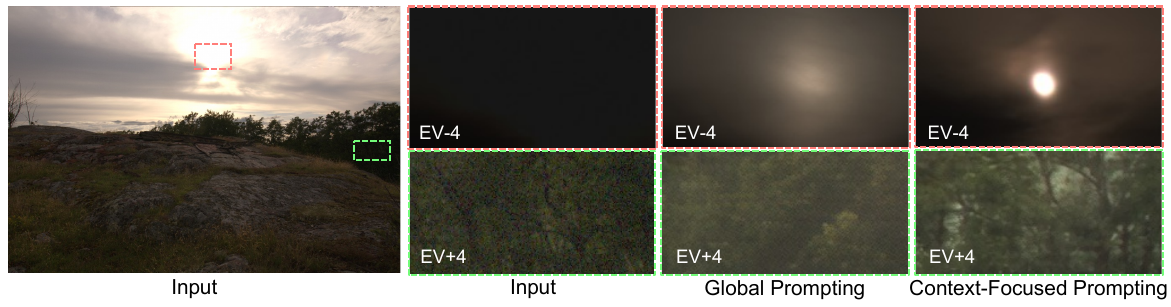}
    \vspace{-4ex}
    \caption{\textbf{Ablation study on context-focused prompting.} We compare our context-focused prompting with global prompting.}
    \vspace{-3ex}
    \label{fig:ablation_cfa}
\end{figure*}

\begin{figure*}[!t]
    \centering
    \includegraphics[width=\linewidth]{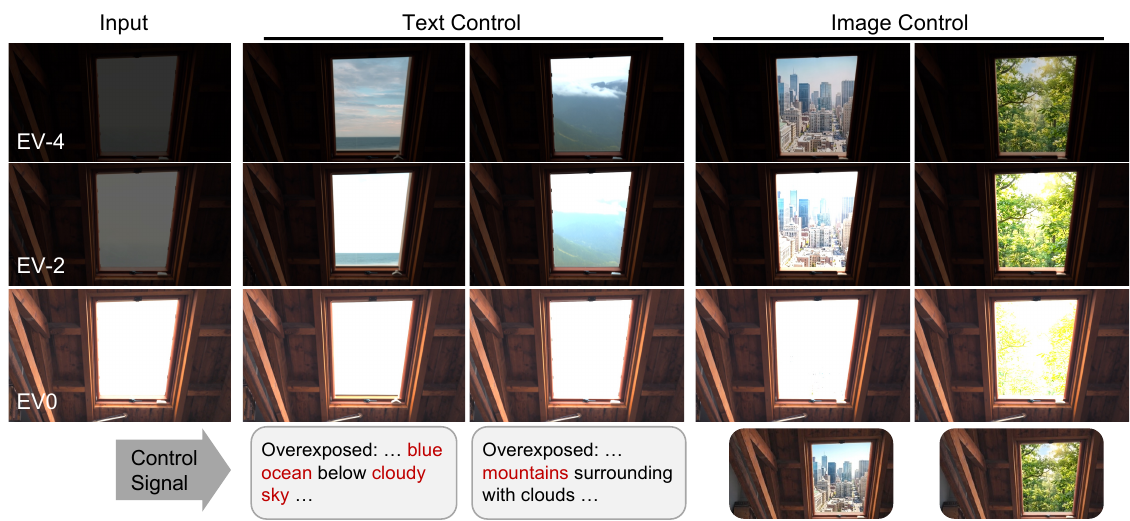}
    \vspace{-2ex}
    \caption{\textbf{Text- and image-guided generation.} DiffHDR supports both text and image controls for guiding the generation in reconstructed regions.}
    \vspace{-3ex}
    \label{fig:control}
\end{figure*}

These observations are further supported by the quantitative results in Tab.~\ref{tab:ablation_encode}. For metric computation, reconstructed HDR outputs are converted to sRGB space and compared against the ground-truth sRGB images using PSNR, SSIM, and LPIPS. Our Log-Gamma mapping achieves the best performance, demonstrating its compatibility with the pretrained VAE.
\vspace{-1ex}
\paragraph{Effect of data augmentation and mask guidance.}
As shown in Fig.~\ref{fig:ablation_aug_mask}, removing our exposure-aware data augmentation during training leads to insufficient noise suppression, resulting in visibly noisy outputs. Without mask guidance, the model struggles to correctly inpaint shadow textures. In contrast, the full model effectively suppresses camera noise and restores detailed textures. These improvements are quantitatively validated in Tab.~\ref{tab:ablation_data_aug}.

\vspace{-1ex}

\paragraph{Effect of context-focused prompting.}
We further evaluate the proposed context-focused prompting modules in Fig.~\ref{fig:ablation_cfa}. When using global prompts alone, the model fails to reconstruct accurate high-intensity structures, such as the sun’s shape. With context-focused prompting, the model successfully restores the correct solar structure and shadowed tree regions.

\begin{wraptable}{r}{6cm}
\vspace{-7ex}
\centering
\caption{Ablation study on data augmentation and mask guidance.}
\label{tab:ablation_data_aug}
\resizebox{0.48\textwidth}{!}{
\begin{tabular}{lcccc}
\toprule
Method & FOVVDP $\uparrow$ & DOVER $\uparrow$ & MUSIQ $\uparrow$ & CLIPIQA $\uparrow$ \\
\midrule
Ours w/o data aug & 7.57 & 0.67 & 59.86 & 0.49 \\
Ours w/o mask     & 7.58 & 0.67 & 59.42 & 0.48 \\
\textbf{Ours}     & \textbf{7.65} & \textbf{0.68} & \textbf{60.02} & \textbf{0.50} \\
\bottomrule
\end{tabular}
}
\vspace{-4ex}
\end{wraptable}

\subsubsection{Controllable Generation.}
In many real-world LDR videos, severely saturated regions may correspond to multiple plausible underlying radiance configurations, leading to inherent ambiguity in HDR reconstruction. To leverage the generative capability of the video diffusion model, our framework enables controllable HDR reconstruction guided by text prompts or reference images.

As shown in Fig.~\ref{fig:control}, by providing different textual descriptions or image references, \diffhdr generates diverse and semantically consistent HDR outputs within overexposed regions. The reconstructed radiance not only aligns with the conditioning inputs but also remains temporally coherent across frames. These results demonstrate that our method goes beyond deterministic restoration and supports controllable, content-aware HDR video synthesis.

\section{Conclusion}
\label{sec:conclusion}

We presented \diffhdr, a novel video diffusion-based framework for generative LDR-to-HDR reconstruction. By reformulating conversion as a radiance inpainting problem within the latent space of a pretrained video diffusion model, our approach leverages strong spatiotemporal priors to recover plausible HDR radiance in overexposed and underexposed regions while maintaining temporal coherence. The proposed Log-Gamma color mapping enables HDR modeling without modifying the pretrained VAE, effectively bridging the distribution gap between LDR and HDR videos. Combined with our HDR video curation pipeline and exposure-aware control mechanisms, \diffhdr achieves state-of-the-art performance across synthetic and real-world benchmarks. Our framework further supports controllable HDR reconstruction guided by text or reference images, opening new possibilities for creative post-production. This work establishes a promising direction for integrating generative video models into practical HDR reconstruction and re-exposure workflows.


%
%

\clearpage
\bibliographystyle{splncs04}
\bibliography{main}

\clearpage
\appendix

\vspace{1cm}
\begin{center}
\textbf{\LARGE Appendix} 
\end{center}

\section{More Results}
\begin{figure*}
    \centering
    \includegraphics[width=\linewidth]{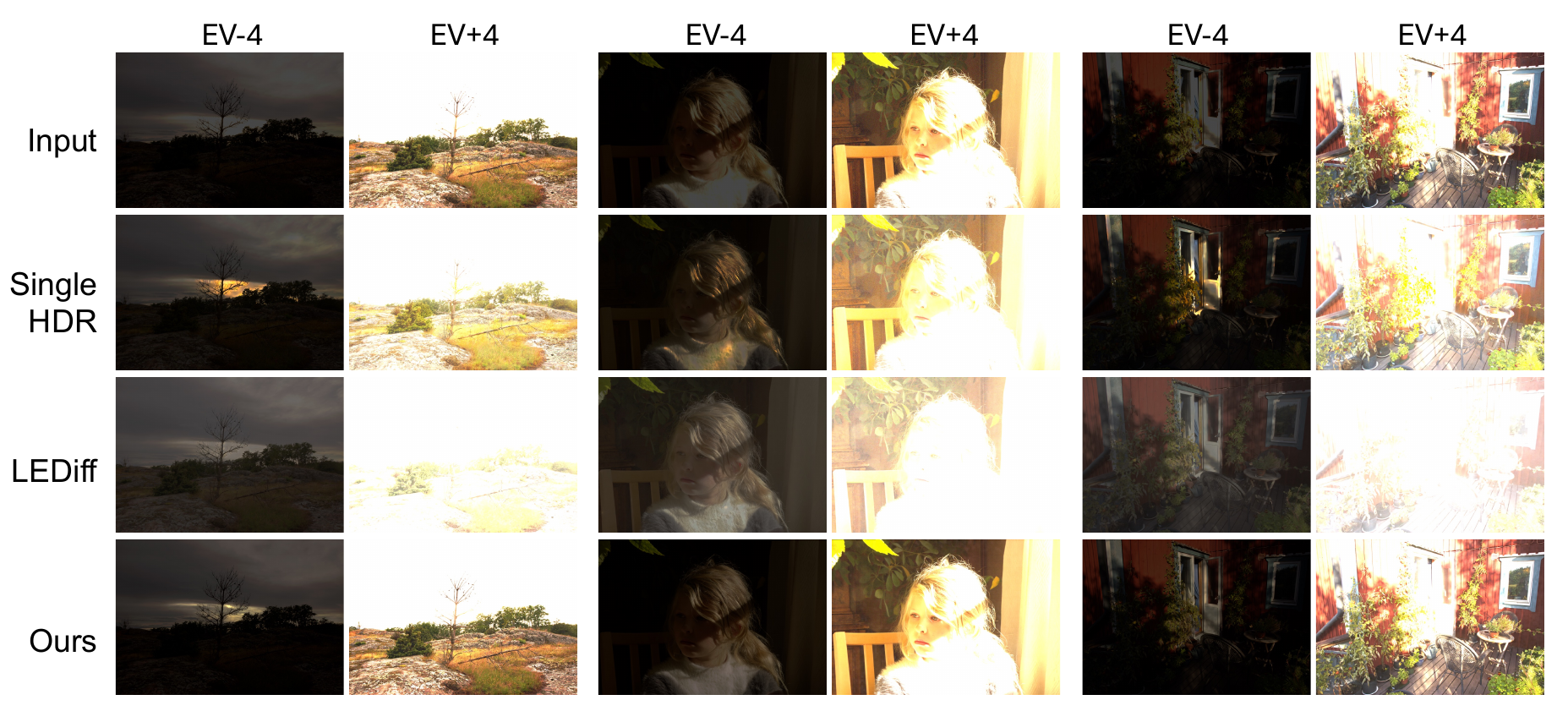}
    \vspace{-3ex}
    \caption{\textbf{Qualitative comparison on the SI-HDR dataset.} 
Results are shown under multiple re-exposure levels. Zoom in for detailed comparison.}
    \label{fig:supp_qual_sihdr}
\end{figure*}

\begin{figure*}[!t]
    \centering
    \includegraphics[width=\linewidth]{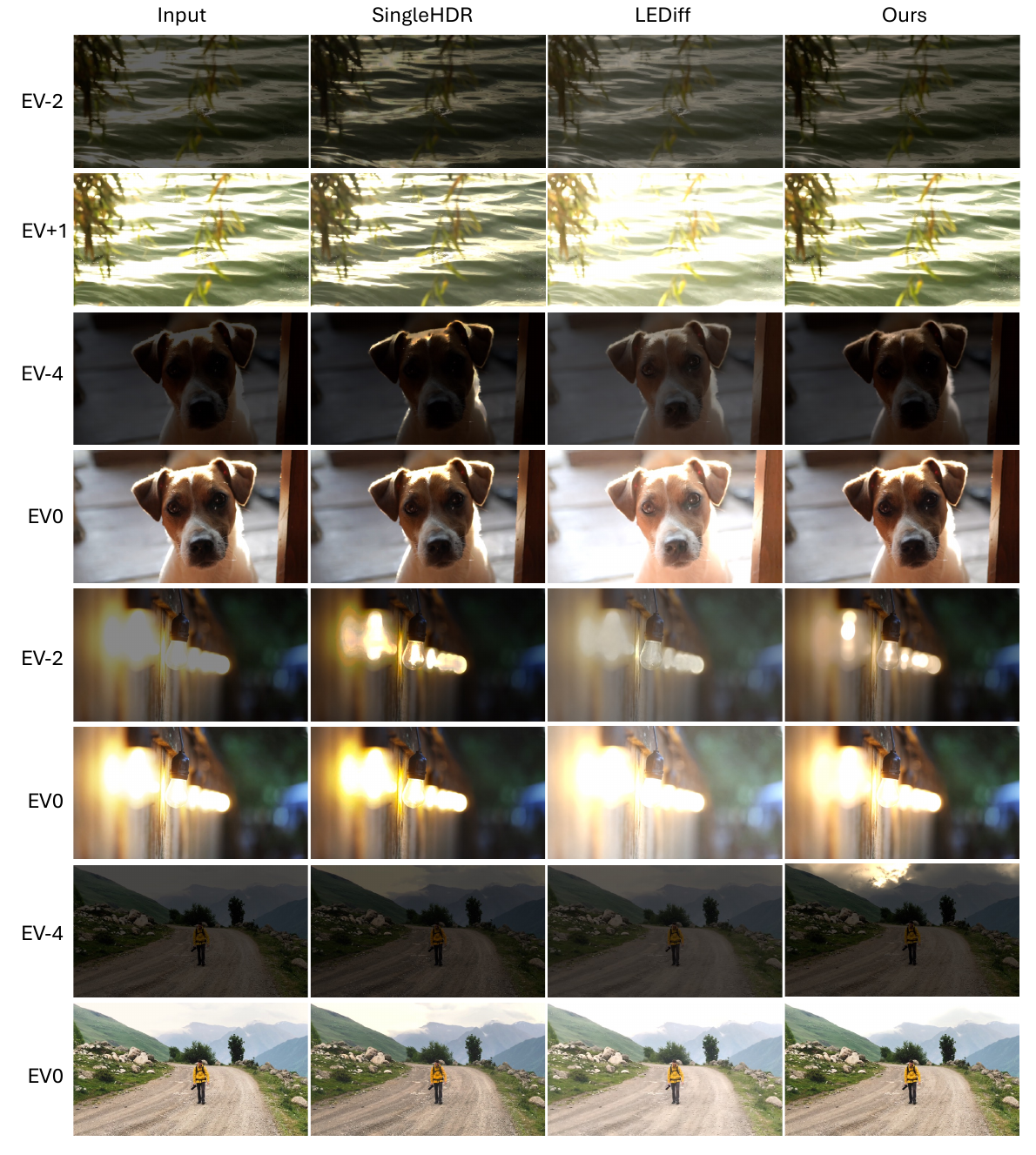}
    \vspace{-3ex}
    \caption{\textbf{Qualitative comparison on the in-the-wild videos.} 
Results are shown under multiple re-exposure levels. Zoom in for detailed comparison.}
    \label{fig:supp_qual_video}
    \vspace{-3ex}
\end{figure*}

We provide additional qualitative comparisons to further demonstrate the effectiveness of \diffhdr. Figure~\ref{fig:supp_qual_sihdr} presents more examples on the SI-HDR~\cite{hanji2022comparison} dataset, including scenes with severe highlight saturation. As shown in these cases, \diffhdr effectively restores fine details in highly saturated regions and preserves structural fidelity when re-exposed to higher exposure levels. In contrast, existing methods often fail to reconstruct plausible content in saturated areas and tend to lose shadow structures under higher exposure due to their limited dynamic range. We further present additional results on in-the-wild videos in Fig.~\ref{fig:supp_qual_video}. Across diverse scenes with challenging illumination conditions, \diffhdr successfully restores saturated regions while producing a wider dynamic range and more faithful scene reconstruction.

\label{sec:experiments}

\section{VAE Finetune Analysis}
\begin{figure*}[!t]
    \centering
    \includegraphics[width=\linewidth]{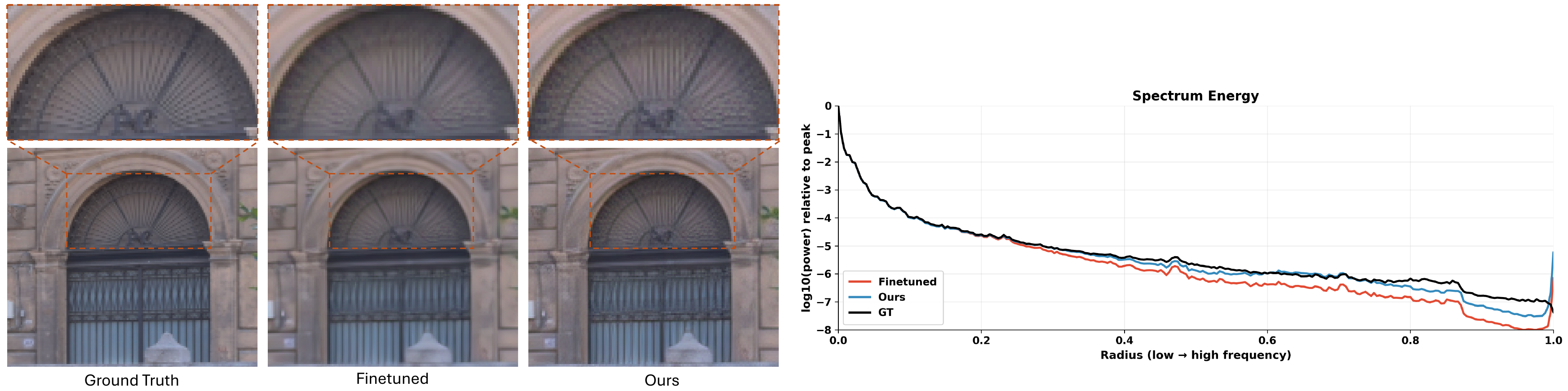}
\caption{\textbf{Effect of finetuning the video VAE.}
The finetuned VAE produces smoother reconstructions and suppresses high-frequency details. The spectrum analysis on the right (radially averaged power spectrum) shows reduced high-frequency energy after finetuning, indicating over-smoothing of the representation. In contrast, the non-finetuned VAE preserves more high-frequency information and yields better reconstruction quality.}
\vspace{1cm}
    \label{fig:supp_vae}
\end{figure*}

To assess whether finetuning the Video VAE improves the encoding and decoding of HDR content, we compare models with and without VAE finetuning by evaluating reconstructed HDR frames (obtained by encoding and then decoding the input). The VAE is finetuned on our HDR video dataset using a standard VAE training procedure. As shown in Fig.~\ref{fig:supp_vae}, the finetuned VAE produces noticeably smoother reconstructions, indicating that high-frequency structures are attenuated during encoding-decoding process. On the right, we visualize the spectrum energy (radially averaged power spectrum over spatial frequencies), which further confirms that finetuning reduces high-frequency energy compared to the non-finetuned baseline. This suggests that VAE finetuning tends to over-smooth the latent representation and suppress fine details that are beneficial for downstream generation. In contrast, the non-finetuned Video VAE preserves more high-frequency information and achieves better overall performance, making additional VAE finetuning unnecessary in our setting.

\section{Banding Effects in VAE}

\begin{figure*}[t]
    \centering
    \includegraphics[width=\linewidth]{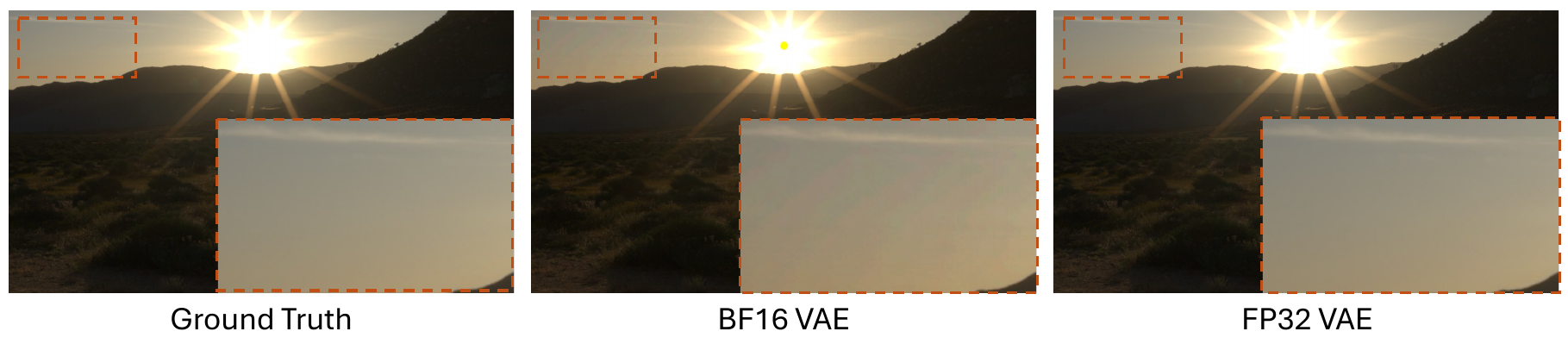}
    \caption{\textbf{Banding artifacts caused by BF16 inference in the video VAE.}
    When decoding HDR images with BF16 precision, the VAE produces visible banding artifacts in smooth intensity regions. In contrast, FP32 inference eliminates these artifacts and yields more realistic reconstruction.}
    \label{fig:supp_banding}
    \end{figure*}

Using video VAE to decode HDR content with BF16 precision can introduce banding artifacts due to its limited numerical precision. 
These artifacts mainly appear in dark regions with smooth luminance gradients (see Fig.~\ref{fig:supp_banding}). In contrast, FP32 inference provides substantially higher numerical precision, enabling finer representation of intensity variations and effectively eliminating these artifacts. Therefore, we use BF16 to finetune the DiT while maintaining the VAE in FP32 to preserve reconstruction quality.

\label{sec:experiments}

\section{Effect of alpha in Context-Focused Cross-Attention}
\begin{figure*}[t]
    \centering
    \includegraphics[width=\linewidth]{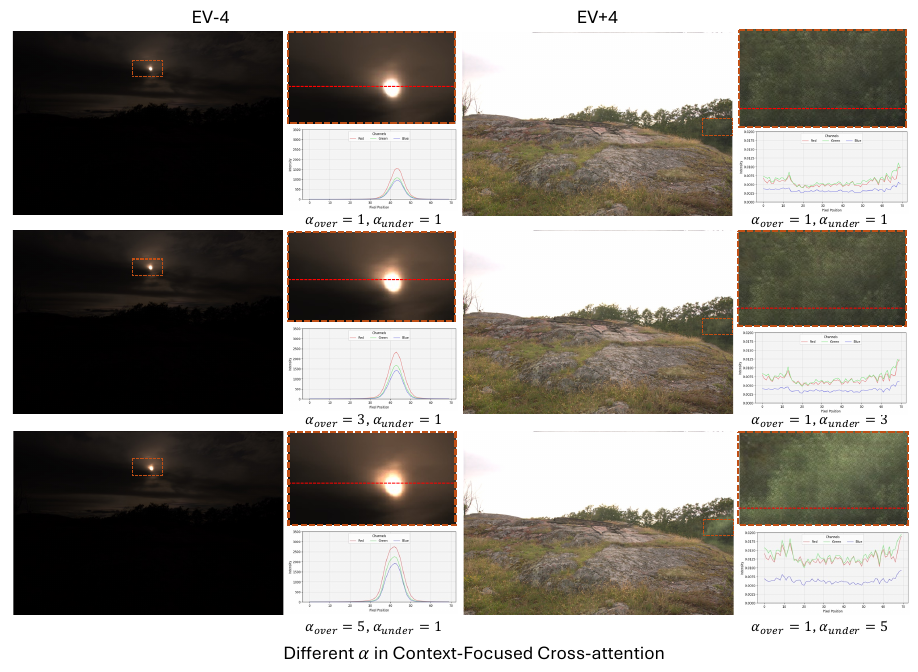}
\caption{\textbf{Effect of $\alpha$ in context-focused cross-attention.}
By adjusting the control coefficient $\alpha$, our method enables controllable manipulation of the dynamic range in overexposed and underexposed regions. As $\alpha$ increases, the recovered radiance in the corresponding regions becomes progressively stronger, leading to larger dynamic range as illustrated by the intensity visualization.}
    \label{fig:supp_alpha}
\end{figure*}

By adjusting the control coefficients (e.g., $\alpha$) for overexposed and underexposed regions, our method enables effective manipulation of dynamic range in the corresponding areas. As shown in Fig.~\ref{fig:supp_alpha}, the recovered dynamic range increases with respect to the associated control parameters shown with the intensity figure, demonstrating controllable HDR reconstruction.

\section{Data Captioning}

To obtain semantic supervision for training, we automatically generate text descriptions for our video data using the Qwen3-VL~\cite{yang2025qwen3} vision-language model. Since our dataset consists of HDR videos stored in linear radiance space, the raw HDR frames cannot be directly processed by standard vision-language models that are trained primarily on LDR imagery. 

Therefore, before caption generation, we first convert the HDR frames into LDR images using Reinhard tonemapping~\cite{reinhard2005dynamic}. This step compresses the dynamic range while preserving the overall scene structure and visual semantics, enabling reliable caption generation. For each video clip, we uniformly sample representative frames and apply the Reinhard tone-mapping operator to convert them into displayable LDR images.

These processed frames are then fed into the Qwen3-VL model to generate textual descriptions of the scene content. The generated captions focus on the overall scene layout, objects, and environmental context, which provide semantic guidance during training. In practice, we use a structured caption format that explicitly separates regions with different exposure characteristics. Specifically, the generated descriptions follow the format:
\texttt{[Overexposed: <description>]; [Underexposed: <description>]}.
This representation allows the model to better associate semantic cues with regions affected by highlight saturation or shadow noise. The resulting captions are used as conditioning inputs for training the HDR reconstruction model.

\label{sec:experiments}

\label{sec:experiments}

\label{sec:experiments}

\end{document}